# Deep learning for predicting hauling fleet production capacity under uncertainties in open pit mines using real and simulated data

N Guerin[1], M Nakhla[2], A Dehoux[3] and J L Loyer[4]

1. PhD student, Mines Paris – PSL, Paris 75006, France. Email: nicolas.guerin@minesparis.psl.eu
2. Professor, Mines Paris – PSL, Paris 75006, France. Email: michel.nakhla@minesparis.psl.eu
3. Lead Data Scientist, Eramet, Paris 75015, France. Email: anita.dehoux@eramet.com
4. Head of Data and AI, Eramet, Paris 75015, France. Email: jl.loyer@eramet.com

## ABSTRACT

Accurate short-term forecasting of hauling-fleet capacity is crucial in open-pit mining, where weather fluctuations, mechanical breakdowns, and variable crew availability introduce significant operational uncertainties. We propose a deep-learning framework that blends real-world operational records (high-resolution rainfall measurements, fleet performance telemetry) with synthetically generated mechanical-breakdown scenarios to enable the model to capture fluctuating high-impact failure events. We evaluate two architectures: an XGBoost regressor achieving a median absolute error (MedAE) of 14.3 per cent and a Long Short-Term Memory network with a MedAE of 15.1 per cent. Shapley Additive exPlanations (SHAP) value analyses identify cumulative rainfall, historical payload trends, and simulated breakdown frequencies as dominant predictors. Integration of simulated breakdown data and shift-planning features notably reduces prediction volatility. Future work will further integrate maintenance-scheduling indicators (Mean Time Between Failures, Mean Time to Repair), detailed human resource data (operator absenteeism, crew efficiency metrics), blast event scheduling, and other operational constraints to enhance forecast robustness and adaptability. This hybrid modelling approach offers a comprehensive decision-support tool for proactive, data-driven fleet management under dynamically uncertain conditions.

## INTRODUCTION

The increasing competitiveness and constant economic pressure are compelling the mining industry to rigorously optimise operational efficiency and the accuracy of production planning. Traditionally, value engineering and approaches based on Value Driver Trees (VDT) have enabled operators to identify key levers for improvement and prioritise performance indicators (Miles, 1961; Cambitsis, 2012). However, these methods struggle to accurately quantify the impacts of numerous uncertainties, such as unexpected operational disruptions or sudden weather changes, particularly on short-term production plans (Asif, Bessant and Francis, 2010; Carvalho *et al*, 2019; Sánchez *et al*, 2020).

Effectively accounting for these uncertainties is critical, as they directly influence the performance of haulage fleets in open pit mines, where even minor fluctuations can cause significant delays and additional operational costs. Studies have shown that extreme weather conditions, such as heavy rainfall, can lead to substantial annual production losses, reaching up to 10 per cent in some open pit mines. Furthermore, unforeseen equipment failures represent a major source of downtime and productivity loss.

In light of these limitations, advanced machine learning approaches, particularly Deep Learning, have emerged as promising alternatives to improve forecasting accuracy. Recently, several studies have begun exploring these possibilities. For instance, Baek and Choi (2020) employed deep neural networks to forecast ore production based on truck haulage operational data. Fan *et al* (2022), on the other hand, combined tree-based ensemble models with Gaussian mixture models to predict truck productivity, focusing primarily on internal mining operation variables.

However, these studies generally focus on limited sets of internal operational factors and fail to sufficiently capture external dynamics such as detailed weather conditions, which are nonetheless critical for operational reliability. Our study distinguishes itself by explicitly incorporating these external meteorological variables using historical rainfall data, and by leveraging both real and simulated data to enrich the predictive models. The inclusion of simulated data, generated using methods such as Monte Carlo Tree Search (MCTS), specifically enables the analysis of rare but high-impact scenarios such as sudden mechanical failures or significant fluctuations in resource usage (Browne *et al,* 2012).

To achieve this, we propose the use of two complementary models: the XGBoost model to clearly identify key and interpretable features with strong influence on productivity, and a Long Short-Term Memory (LSTM) network capable of capturing the complex temporal dependencies inherent in sequential operational data. The main advantage of our approach lies in this combination of interpretability and predictive performance, aiming to provide a practical and directly usable tool for operational managers to anticipate and adjust their decisions in real time in the face of uncertainties.

This research directly addresses a need in the mining industry: enhancing operational resilience and improving proactive fleet management. By integrating an advanced analytical perspective and a better understanding of uncertainty dynamics, we lay the foundation for more robust operational management, with the potential to have a direct impact on the economic performance of modern mining operations.

## MATERIALS AND METHODS

The following section outlines the foundational components of our study: the data sourced from a mining fleet management system (FMS) and framework designed to predict hauling fleet productivity. In production systems, the performance depends on both operator efficiency and equipment availability. However, due to the lack of human resources data in our study, we have chosen to focus solely on the trucks as our primary resource. This approach enables us to concentrate our analysis on the fleet's impact on operational capacity, which is the critical determining factor. By integrating domain-specific data with advanced modelling techniques, this work addresses the unique challenges of mining operations, balancing interpretability and temporal forecasting accuracy.

### Fleet management system

Mining fleet operations generate vast amounts of heterogeneous data, driven by the integration of IoT sensors, GPS modules, and onboard telemetry systems. The data set underpinning this study originates from a fleet of heavy-duty mining equipment operating in a large-scale open pit mine. Several large-scale mining management software packages exist, including HxGN MineOperate, which integrates data communication functions, GPS positioning for shovels, drills, and haul trucks, and automated fleet assignments for open pit mining operations. The system consists of the following key components:

- HxGN MineOperate OP Pro: A computerised field system equipped with an advanced interface, installed on trucks, auxiliary equipment, shovels, and crushers to optimise fleet management and operational workflows.

- Global Positioning System (GPS): Ensuring precise tracking and navigation of mining equipment.

- Radio Data Link: Connected to a central computing hub, facilitating real-time data transmission and enabling automated haul truck assignments through peer-to-peer communication, reducing inefficiencies due to network delays.

The hauling process initiates at the beginning of each shift, where operator data is entered into the system to ensure proper registration of personnel and equipment. This information is transmitted to the central control system, which then enables the dispatcher to assign work to operators. Operators receive real-time notifications regarding equipment utilisation, operational requirements, and mandatory inspections, ensuring continuous workflow efficiency.

The temporal nature of the data provides granular insights into equipment behaviour and operational trends. With over 50 features per equipment unit, the data set combines numerical, categorical, and time-series formats, reflecting the complexity of industrial-scale mining operations.

However, raw industrial data often suffers from inconsistencies inherent to harsh mining environments. To address these issues, preprocessing steps were rigorously applied. Short data gaps were resolved using linear interpolation or forward-fill techniques, while longer gaps were flagged for manual review. García, Luengo and Herrera (2015) provided a comprehensive framework for data preprocessing in complex industrial data sets, advocating for context-aware interpolation strategies to maintain data fidelity

## Models' selection

### *XGBoost for interpretable feature analysis*

XGBoost, a gradient-boosted tree algorithm (Figure 1), was selected for its ability to handle mixed data types, including categorical maintenance codes and numerical sensor readings. Its interpretability is particularly valuable in mining, where stakeholders prioritise understanding the drivers of productivity. Furthermore, its inherent regularisation techniques help mitigate overfitting, which is crucial when dealing with noisy industrial data. In addition, its built-in capabilities to perform embedded feature selection complement the rigorous upstream preprocessing, ensuring robust performance on heterogeneous data sets. For instance, XGBoost's feature importance scores can highlight critical predictors that are subsequently used to inform further feature engineering and operational decision-making. Moreover, the algorithm's scalability and computational efficiency enable rapid model iteration and facilitate real-time deployment in environments with large, continuously streaming data sets.

The efficacy of XGBoost in structured data analysis has been well-documented. Chen and Guestrin (2016) demonstrated its scalability and performance in capturing non-linear relationships, while Wang *et al* (2021) proved its capability to model complex interactions in industrial data. Beyond these applications, our choice is primarily driven by XGBoost's strong theoretical foundations and practical advantages, its ability to naturally capture non-linear feature interactions and its built-in mechanisms to control model complexity make it exceptionally well-suited for our predictive tasks.

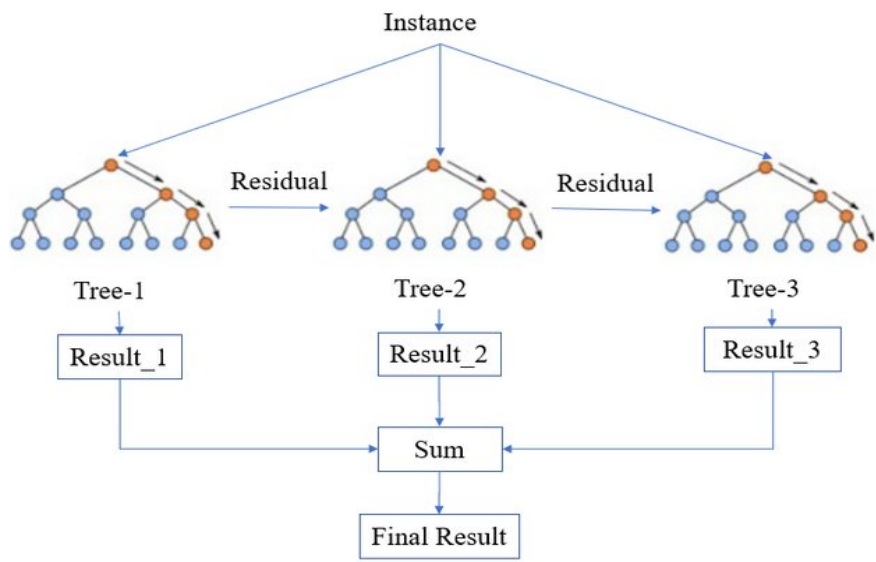

**FIG 1** – XGBoost model scheme from Wang, Chakraborty and Chakraborty (2020).

### *LSTM networks for temporal dynamics*

LSTM networks were implemented to capture the intricate temporal dependencies in equipment behaviour. Their gated architecture (Figure 2) enables effective retention and selective forgetting of information over long sequences. Designed to remember long-term patterns, LSTMs are ideal for mining fleet data, where cyclic production schedules are common. By processing multivariate time-series sequences, the LSTM architecture learns to forecast key metrics, effectively modelling the dynamic evolution of mining operations. Moreover, LSTMs inherently handle variable sequence lengths and are robust to noise and irregular time intervals, common in industrial data sets. This ability to integrate contextual information from previous time steps results in more accurate predictions when past operational conditions heavily influence future outcomes.

The foundational work of Hochreiter and Schmidhuber (1997) established LSTMs as a solution to the vanishing gradient problem, enabling effective learning of long-term dependencies. More recently, Ao, Li and Yang (2023) demonstrated the efficacy of LSTM networks for predicting truck travel time in open pit mines by capturing complex temporal dependencies in historical data, thereby enhancing scheduling and operational efficiency. Our choice is further justified by LSTM's proven performance in handling non-stationarity and noise in time series data, making it an excellent candidate for forecasting in the unpredictable operational environment of open pit mining.

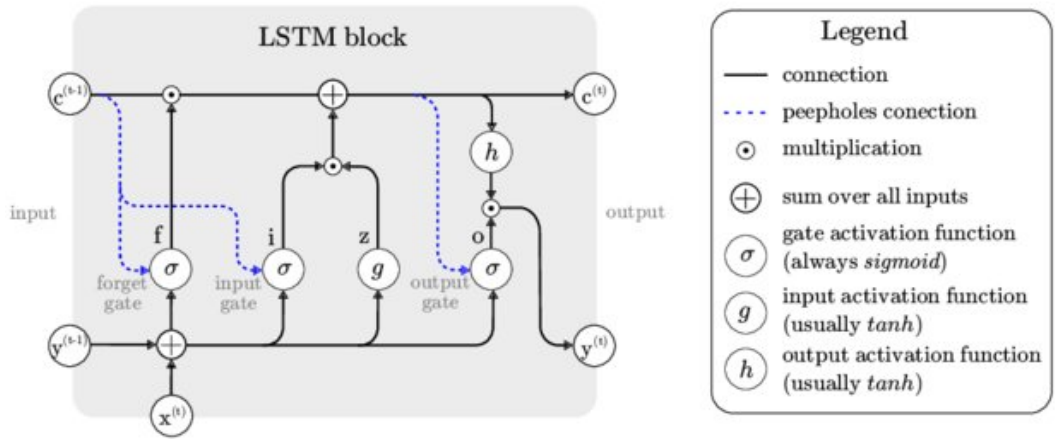

**FIG 2** – LSTM model scheme from Van Houdt, Mosquera and Napoles (2020).

We adopt two distinct modelling approaches to comprehensively analyse mining fleet productivity. The first approach leverages XGBoost, a powerful gradient-boosted tree algorithm, to extract interpretable insights from static, time-windowed data snapshots. The second approach employs LSTM networks, which are adept at capturing the temporal dynamics inherent in sequential equipment behaviour. Together, these methods enable us to dissect both the static and dynamic drivers of operational efficiency in mining fleets.

## ANALYSIS AND PROPOSAL

Currently, large-scale mining operations deploy fleets consisting of hundreds of trucks, dozens of shovels, several loaders, and other equipment distributed across multiple pits. As depicted in Figure 3, these operations are integrated across various mining areas, with certain equipment such as shovels and loaders being flexibly repositioned according to daily operational requirements and mining plans, while other assets remain fixed.

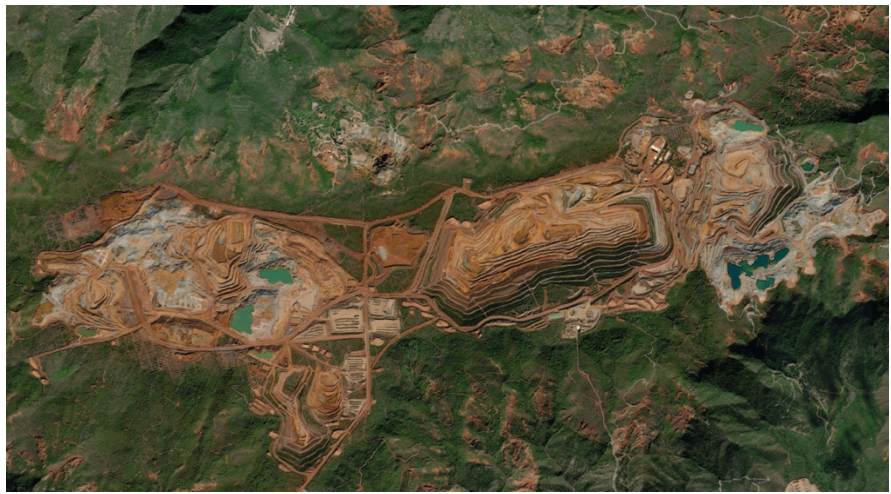

**FIG 3** – Mining operation of two open pits (image from Google Earth).

A central dispatch system manages the allocation of trucks to ensure that their distribution aligns with the overall operational strategy. However, the actual production capacity of the hauling fleet is subject to significant uncertainties. For instance, truck availability can be disrupted by unexpected breakdowns, and adverse weather conditions—particularly rain—can further impact operational efficiency. These factors introduce variability in the effective production capacity, making accurate forecasting a complex challenge.

In our study, we leverage a multi-faceted and temporally rich data set to address these challenges. The data set captures the dynamic nature of mining operations through continuous, time-stamped records shift-by-shift of 10.5 hrs that reflect real-time fluctuations in performance between shifts. Simultaneously, weather data recorded at matching intervals provides insights into environmental conditions that directly influence operations. Additionally, simulated data augments our observations by modelling scenarios critical for understanding changing conditions and their potential impact on production capacity.

By combining these diverse data sources, our approach aims to develop a robust forecasting model that accounts for both the inherent temporal fluctuations in mining operations and the external factors that impact efficiency. This holistic view is essential for improving decision-making in large-scale, multi-pit mining environments.

## Operational data

The operational data utilised in this study are directly sourced from the FMS described in the section earlier. Specifically, the FMS continuously collects detailed, cycle-by-cycle data from individual haul trucks and shovels which are aggregated at the shift level. These data encompass several key performance indicators that reflect the dynamics of our mining operations. Meanwhile, the working trucks metric, which denotes the number of trucks actively engaged in material transport during the shift, averaged 13.8 trucks with a variability of 3.1, while the working shovels feature, indicating the number of shovels in operation, maintained an average of 4.7 with a standard deviation of 1.4, thus ensuring a balanced deployment of equipment. Additionally, the cycle count, signifying the number of complete loading cycles per shift, was observed to average 158 cycles with a standard deviation of 71, directly reflecting the throughput of our operations. The payload, which measures the average weight of material loaded per shift in tons, averaged 13 795 tons with a variability of 6393 tons, underscoring both the efficiency and the inherent variability of material handling.

Furthermore, the average cycle time of a shift, the duration required to complete a full cycle from truck arrival to departure recorded an average of 64 mins with a standard deviation of 17 mins, capturing the relation between operational speed and occasional delays during peak activity. A schematic representation of a hauling truck cycle is provided in figure 4.

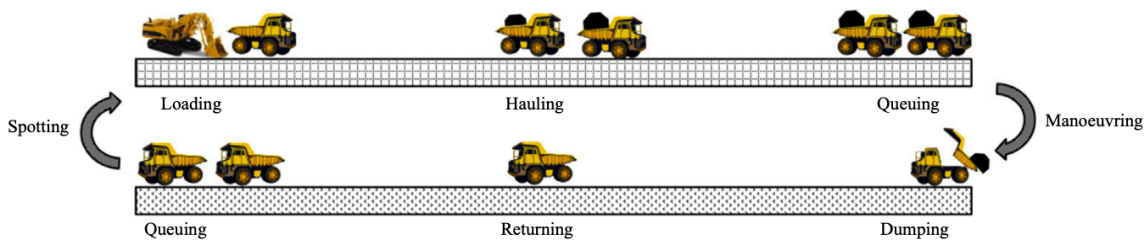

**FIG 4** – Schematic of hauling operation in surface mines from Soofastaei *et al* (2015).

Finally, the shift and crew variables categorise the operational periods, revealing that certain shifts experienced slightly faster cycle times, likely a result of differences in crew performance and working conditions. Collectively, these metrics, quantified through their respective means and standard deviations, provide a comprehensive picture of our operational efficiency and variability, and they form the backbone of our model's input data for predictive analysis and simulation.

## Weather data

Specifically, we leverage the ERA5-Land hourly data set (spanning from 1950 to the present), which is accessible via the Copernicus platform either through direct download or API queries. The data set provides a high spatial resolution of 0.1° × 0.1° and an hourly temporal resolution, ensuring that even short-term variations in weather are captured.

In order to effectively represent the variability in precipitation across the mining area, we extract data from four key cardinal points that delineate the boundaries of the mine. These points, corresponding to the northern, southern, eastern, and western limits of the site, are selected to encompass the full range of local weather influences. Additionally, we selected the grid point closest to the centre of the mine. After collecting the hourly precipitation data from this location, we aggregate the values at the operational post level (Figure 5). This is achieved by summing the hourly measurements over defined time intervals, thereby yielding cumulative precipitation metrics that not only indicate the weather's impact during each operational shift but also capture the intense rainfall peaks that significantly affect operations and reflect seasonal variations.

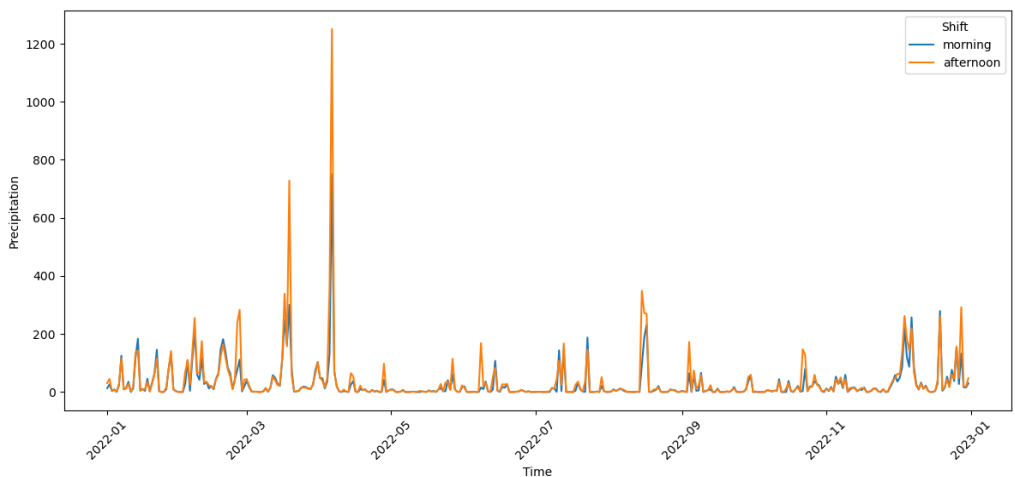

**FIG 5** – Rain precipitations variation on mine over one year (2022).

To illustrate the extent of rainfall variability in our tropical context, we analysed precipitation data region from 2021 to 2024. Annual totals ranged from over 2500 mm in 2021—an exceptionally wet year marked by cyclonic activity—to just 1485 mm in 2024, reflecting a 6 per cent deficit relative to the 1991–2020 baseline. The region displays a clear seasonality, with a wet season spanning January to Mars, during which extreme events such as tropical depressions contribute to intense rainfall peaks. For instance, March 2021 recorded up to 320 mm of rain. In contrast, the dry season from August to November is significantly less active, with monthly rainfall sometimes dropping below 30 mm, as observed in September 2022.

Incorporating cumulative precipitation measures into our deep learning model is crucial, as heavy rainfall episodes directly reduce operational efficiency by deteriorating haul road conditions, increasing cycle times, and occasionally forcing temporary equipment downtime due to unsafe conditions. Gonzalez *et al* (2019) evaluated the effects of extreme rainfall events on open-pit mines in Peru, demonstrating marked operational delays during heavy rains. Similarly, Tlhatlhetji and Kolapo (2021) studied the rainy season at the Wescoal Khanyisa Colliery, documenting a significant decrease in equipment availability and cycle counts during wet periods. Note that we exclusively incorporate rainfall metrics, as temperature in our tropical region lacks significant seasonal variability and thus has been excluded from our analysis.

## Simulated data

Mechanical breakdowns, significantly impact fleet productivity through sudden equipment downtime. Our study incorporates simulated data designed to estimate the potential utilisation of trucks and shovels for upcoming shifts. The objective of this simulation framework is to realistically model the short-term operational capacity of mining fleets using historical availability and utilisation data collected through the FMS.

A detailed statistical analysis of past shifts revealed that truck availability varies significantly from one shift to the next, with an average fluctuation of approximately 22 per cent, corresponding to about 3.1 trucks around a mean of 13.8 per shift. Similarly, shovel utilisation shows an average variation of around 21 per cent, translating to roughly 1.4 shovels around a mean of 4.7 per shift.

These variations, which reflect operational uncertainty, are explicitly embedded into our simulation framework to produce robust and realistic predictions (see Figure 6).

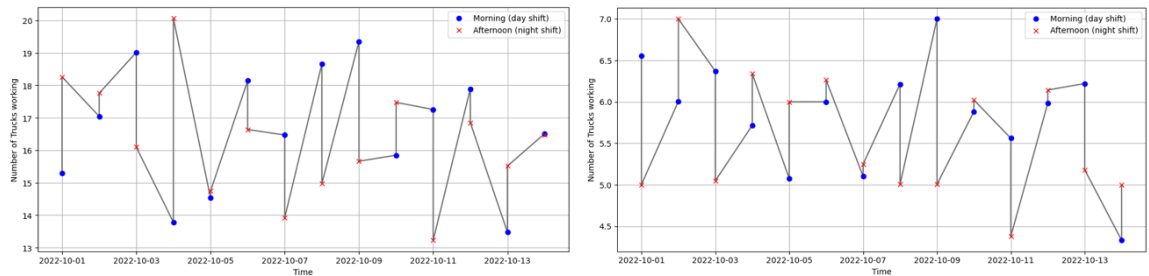

**FIG 6** – Trucks (left) and shovels (right) utilisation variation per shift through time.

To generate these simulations, we adopt a hybrid approach that combines regression modelling with Monte Carlo simulation techniques inspired by MCTS. Instead of relying on fixed historical averages, we first train regression models to predict the fleet composition for the next shift based on empirical data. We then inject stochastic noise sampled from the distribution of model residuals to simulate variability. This process yields a range of plausible future scenarios, providing a nuanced representation of operational uncertainty in mining environments (Browne *et al*, 2012).

## Proposed methodology and models

This research focused on applying time series forecasting learning models, specifically XGBoost and LSTM, within the context of trucks and shovels in mining operations. The research utilised the open-source programming language Python, along with relevant libraries that facilitated the simulation of the mining process and training of the learning models. The methodology employed in this research encompassed a comprehensive analysis of the hauling cycle over a three-year period. Through the utilisation of a Transact-SQL script and the Dispatch system, essential data on trucks and shovels' operational performance, as well as weather, were extracted. A wide range of daily operating scenarios was accounted for. This meticulous methodology ensured a comprehensive and reliable foundation for subsequent learning modelling and simulations.

In addition, we integrated temporal lag features into our modelling framework to capture dynamic dependencies and cumulative effects in mining operational data. Each feature was designed to provide specific insights into the system's behaviour. For example, we created two payload-related temporal lags: one that represents the payload recorded during the previous time period—capturing short-term persistence in payload delivery—and another that computes the rolling sum of payloads over the last four shifts, which smooths out short-term fluctuations and highlights sustained trends. Similarly, two lag features related to fleet operations were implemented: one indicates the number of working trucks in the immediate past period, reflecting the current operational state, while the other aggregates the average number of working trucks over four shifts, offering a broader view of fleet performance over time. Finally, we computed the cumulative sum of rainfall over the last six shifts to quantify the overall impact of weather events on operations and road conditions. By incorporating these features, our model leverages both immediate past values and aggregated historical trends, thereby enhancing its ability to predict future operational performance accurately.

Incorporating these temporal lag features is essential in time series analysis, as highlighted in the literature (Hyndman and Athanasopoulos, 2018). They not only enrich the feature set by embedding historical context but also enhance the predictive capabilities of the models by capturing both immediate and aggregated temporal dynamics.

To ensure transparency and reproducibility, Table 1 summarises the input features used in both models, clearly indicating their data sources, nature (empirical, simulated, predictive, temporal lag), and precise definitions. These features were selected based on their operational relevance and predictive value.

**TABLE 1**

Model input and metadata.

| Feature name | Data source | Feature type | Description |
|---|---|---|---|
| Crew_next | Operational | Predictive | Crew scheduled for the next shift |
| Working_trucks | Operational | Empirical | Number of trucks operating during the current shift |
| Predicted_working_trucks_next | Simulated | Predictive | Predicted number of trucks for the next shift |
| Predicted_working_shovels_next | Simulated | Predictive | Predicted number of shovels for the next shift |
| Working_shovels | Operational | Empirical | Number of shovels operating during the current shift |
| Cycle_count | Operational | Empirical | Number of loading cycles completed per shift |
| Payload | Operational | Empirical | Total payload transported during the shift (in tons) |
| Cycle_time | Operational | Empirical | Average duration of loading cycles per shift |
| Payload_lag1 | Operational | Temporal Lag | Payload from the previous shift |
| Payload_rolling_sum_4 | Operational | Temporal Lag | Sum of payloads over the last four shifts |
| Shift_next | Operational | Predictive | Next shift scheduled (day or night) |
| Working_trucks_lag1 | Operational | Temporal Lag | Number of trucks operating during the previous shift |
| Working_trucks_mean4 | Operational | Temporal Lag | Mean number of working trucks over the last four shifts |
| Precipitation | Meteorological | Empirical | Total rainfall during the previous shift |
| Precipitation_next | Meteorological | Predictive | Historical rainfall for the next shift |
| Precipitation_sum6 | Meteorological | Temporal Lag | Cumulative rainfall over the last six shifts |

In this study, two distinct forecasting models were developed to predict the next payload in mining operations: an XGBoost regression model and a Long Short-Term Memory (LSTM) network.

We began by extracting the relevant features from our data set and normalising them using a MinMaxScaler. To preserve the time-dependent structure, the data was split chronologically (80 per cent for training and 20 per cent for testing). The XGBoost regressor was then configured with hyperparameters selected to balance model complexity and generalisation. In particular, we set:

- ***n_estimators***: 1000
- ***Learning Rate:*** 0.01
- ***Max Depth:*** 3

After training on the training subset and evaluating on the test subset, we conducted a feature importance analysis. To further unpack variable interactions, Shapley Additive Explanations (SHAP) values were computed, offering granular insights into how specific features influence predictions.

In contrast, the LSTM model was tailored to harness the sequential characteristics of our operational data. After removing missing values and scaling both the features and target variable, a sliding window approach with a look-back period of ten-time steps was used to create temporal sequences. The LSTM architecture featured an LSTM layer with 64 units to process these sequences, followed by a dropout layer (with a dropout rate of 0.2) to mitigate overfitting, and a dense layer to output the final prediction. Moreover, to enhance robustness, an attention mechanism was integrated to highlight critical temporal intervals—such as periods of peak engine stress—that could significantly impact performance. The model was compiled using the Adam optimiser (with a learning rate of 0.001 and a clip value of 0.5) and trained using mean squared error as the loss function, with early stopping employed to capture the best model state.

After training, predictions were made on the test set, and the outputs were inverse transformed to their original scale. Performance metrics (MedAE and $R^2$) were computed for both models, allowing us to draw a direct comparison.

## RESULTS

In our study, two forecasting models were developed to predict payload in mining operations. To assess their performance, we selected the median absolute error (MedAE) as our primary evaluation metric due to its robustness against outliers and its interpretability in the context of operational variability.

One of the key innovations in our approach was the integration of rainfall data as a predictor. Rainfall is known to affect operational efficiency, and its inclusion allowed the models to capture weather-induced variations in payload. In the XGBoost model, we computed SHAP values to dissect the influence of each feature. The SHAP analysis revealed that rainfall was among the most significant predictors, with increased rainfall correlating with decreased payload performance as shown in Figure 7. This granular insight not only validates our feature selection but also underscores the practical importance of including weather data in operational forecasts.

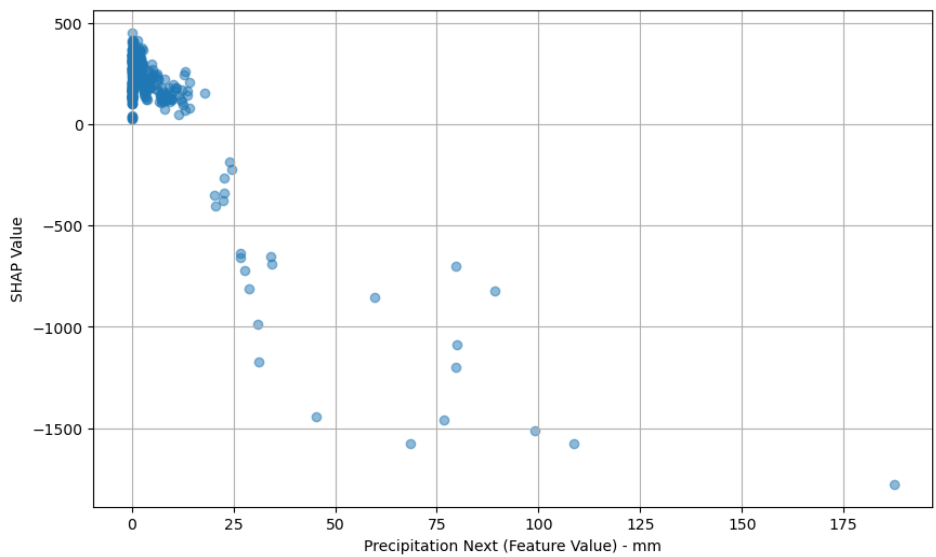

**FIG 7** – SHAP analysis: 'upcoming rainfall' feature impact on model predictions.

To better understand the impact of each input feature on the model's predictions, a sensitivity analysis was conducted using SHAP. Figure 8 presents the SHAP summary plot, where each point represents a prediction, and each colour encodes the feature value. The horizontal spread indicates the magnitude and direction of the feature's impact on the output variable.

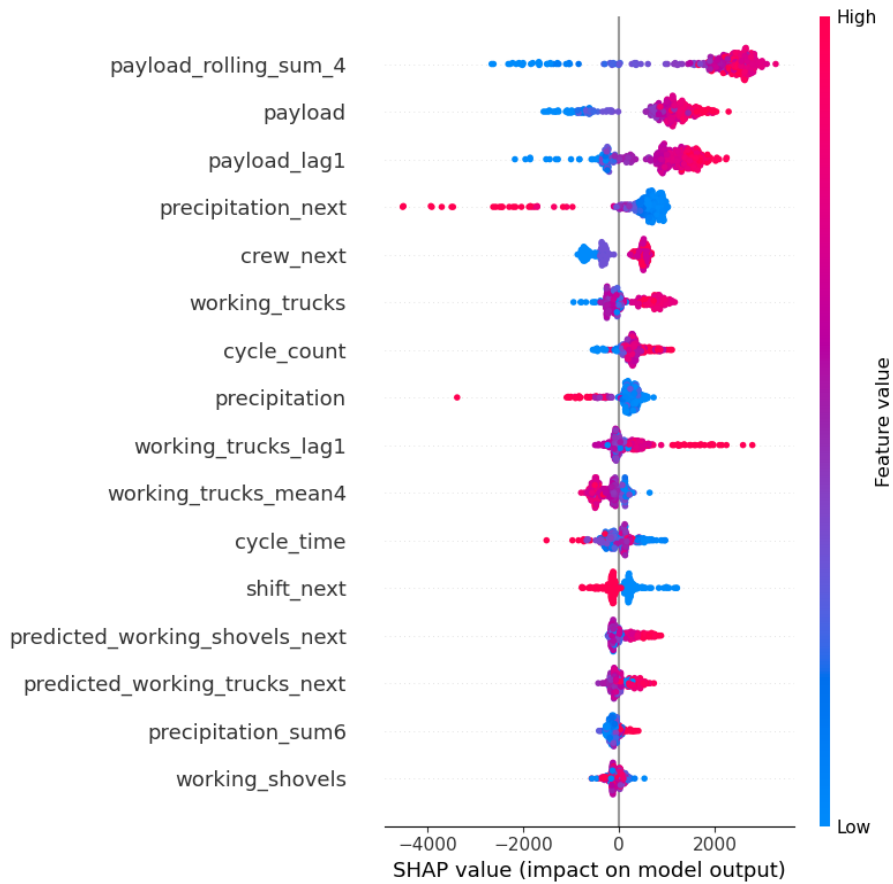

**FIG 8** – SHAP summary plot showing the impact of each feature on model output.

The three variables related to historical payload clearly dominate in terms of impact on the model. The observed trend is intuitive: a low payload in previous periods strongly contributes to predicting a low upcoming payload, while a high previous payload tends to indicate a high subsequent level, illustrating strong operational inertia across successive shifts.

Regarding precipitation, as previously discussed, heavy forecasted rainfall has a significant negative impact on productivity, more so than accumulated past rainfall.

We also observe that the variable representing the next shift has a moderate but clear impact. This directly reflects field experience, where the median payload observed during the night is often higher than during the day, mainly due to fewer ancillary activities at night (eg fewer interruptions or preventive maintenance).

Simulated variables show a relatively weak impact, suggesting that integrating maintenance data and HR information (eg actual availability of operators and equipment) would be necessary to significantly improve the predictive relevance of these indicators. Currently, the impact of actual observed availability during the previous shift outweighs that of simulated forecasts, highlighting the room for improvement in integrating real-time information into these predictive variables.

Finally, the next crew to operate has a moderate but structured impact. This regularity highlights its strategic role in understanding overall operational dynamics.

To visualise model performance, we generated several time-series plots comparing the actual payload with the predicted values from each model over a representative period.

For example, the time series analysis presented in Figure 9 compares the actual and predicted payload values over the last 100 shifts of our data set. The blue line represents the observed payload, while the orange line corresponds to the predictions generated by the model, XGBoost and LSTM respectively.

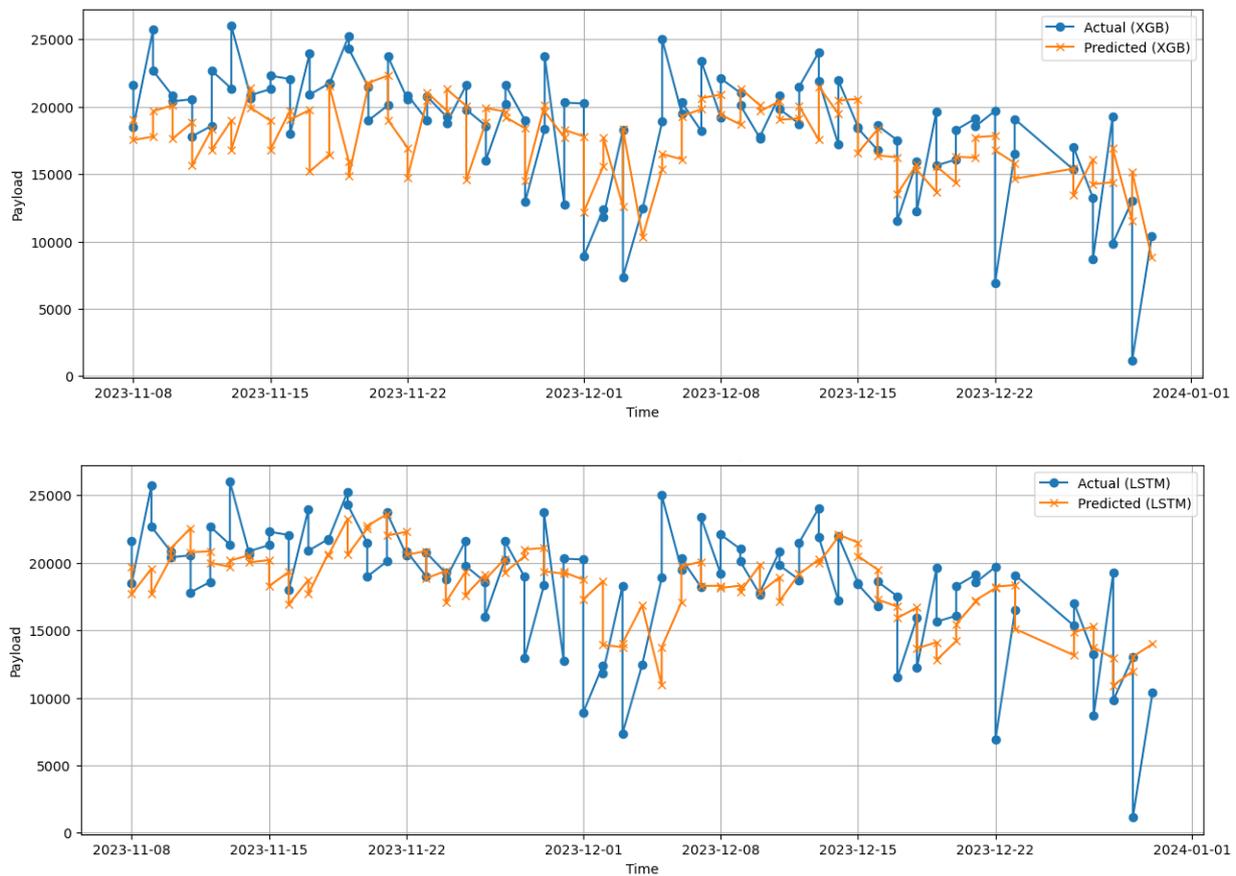

**FIG 9** – XGBoost and LSTM next payload forecast respectively versus actual observations over time.

Both models successfully capture the general trend in the data, but they exhibit different behaviours regarding fluctuations and short-term variations. In the current stage of the work, particular attention is being given to the test data to evaluate the robustness and generalisation capabilities of the models. The XGBoost model, being tree-based, shows a greater ability to react to sharp increases and decreases in payload size. This is particularly useful in scenarios where sudden changes occur frequently, as the model can adapt quickly to new patterns. However, this reactivity can sometimes lead to overfitting, where the model becomes too sensitive to noise and short-term variations, potentially reducing its generalisation ability.

In contrast, the LSTM model, due to its sequential learning nature, produces smoother predictions. This stability makes it effective for long-term forecasting and general trend recognition, but it struggles with rapid variations in the payload. The LSTM model tends to exhibit a lagging effect, where it adjusts to changes more slowly than XGBoost. As a result, it is less prone to overfitting but may fail to capture important local fluctuations that influence the overall prediction quality.

Throughout the observed period, both models follow the overall downward trend in payload values towards the end of December 2023. However, the XGBoost model better anticipates sudden peaks and drops, whereas the LSTM model exhibits delay in adapting to these changes. This suggests that while LSTM is better at capturing the broader structure of the time series, it lacks the flexibility needed to track short-term anomalies effectively.

To further evaluate model performance, the percentage error of predictions was analysed for both models over 379 test observations. The evolution of these errors is presented in Figure 10.

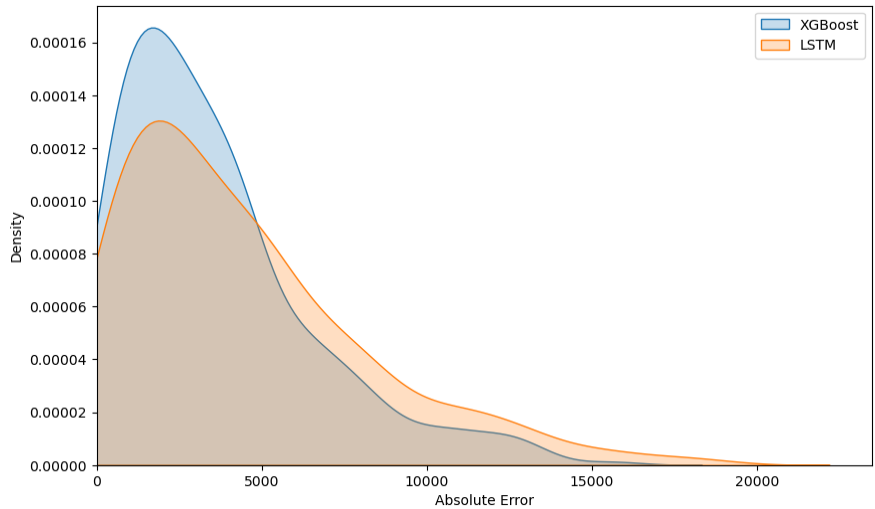

**FIG 10** – Kernel Density Estimate (KDE) of Absolute Errors for XGBoost versus LSTM.

**XGBoost**: Achieved a **MedAE of 14.3 per cent**. This model benefits from its capacity to model non-linear relationships and its enhanced interpretability via SHAP values, which clearly indicate the strong influence of rainfall on predictions.

**LSTM**: Recorded a **MedAE of 15.1 per cent**. While the LSTM model is particularly adept at capturing sequential dependencies and overall temporal patterns, it occasionally produces larger errors during abrupt operational shifts—often coinciding with heavy impacts events.

The analysis reveals the following key insights: XGBoost produced 33 instances where the error exceeded 50 per cent between the predicted and actual values. LSTM showed a slightly higher count, with 31 instances surpassing the 50 per cent error threshold.

As show in Figure 11 These errors occurred at similar points in time, indicating that both models struggled under the same conditions. A deeper investigation of these high-error occurrences reveals that they do not always coincide with identifiable external rainy events, suggesting that environmental factors were not the primary cause. Instead, these errors are likely due to missing or unrepresented information in the data set that the models were not able to account for.

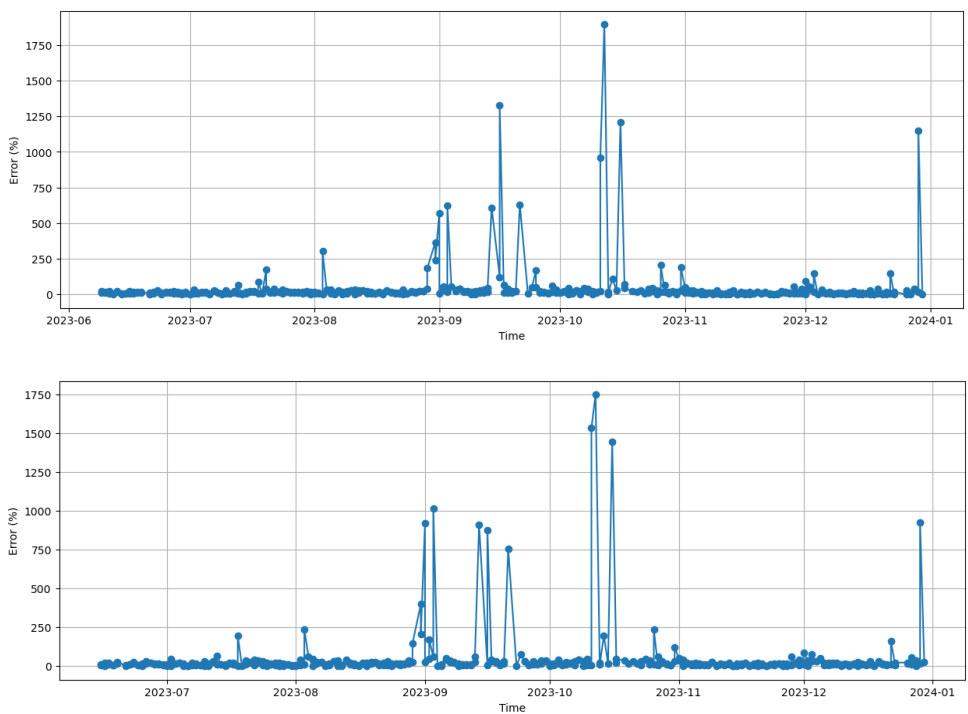

**FIG 11** – Temporal evolution of XGBoost (first) and LSTM (second) forecast error percentage.

A further analysis was conducted to assess the operational impact of these high-error occurrences. We observed that when the number of trucks scheduled for the next shift was incorporated as an input feature, the frequency of large deviations (errors exceeding 50 per cent) dropped significantly—down to just eight instances for XGBoost and 27 for LSTM. This demonstrates that integrating key operational planning metrics has a stabilising effect on XGBoost predictions, significantly reducing high-error occurrences. Furthermore, model performance metrics reinforce these findings:

- XGBoost achieved a **MedAE of 8.4 per cent**, highlighting its robustness in most scenarios.
- LSTM showed a **MedAE of 13.5 per cent**, particularly struggling with lower payload values.
- XGBoost attained an **$R^2$ score of 0.78**, confirming its strong predictive capabilities, as illustrated in Figure 12.

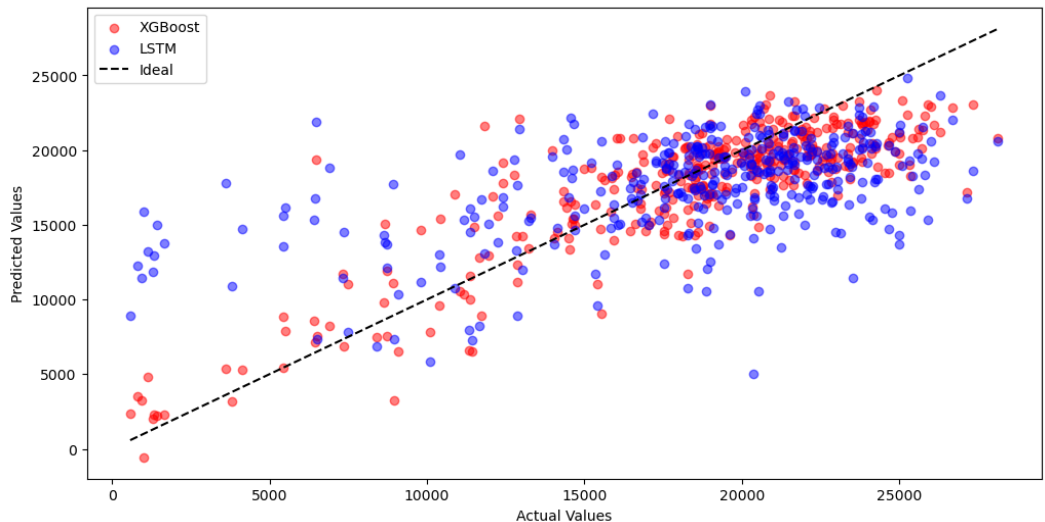

**FIG 12** – Model calibration plot (XGBoost and LSTM).

This analysis underscores the importance of integrating operational factors into predictive modelling. It also opens the door for future enhancements, such as incorporating maintenance scheduling and reliability indicators like Mean Time Between Failures (MTBF) and Mean Time to Repair (MTTR), which could further refine forecasting accuracy and improve overall model stability.

Our results yield several important insights, the clear relationship between rainfall and payload performance, as evidenced by both SHAP values and forecast graphs, underscores the necessity of including weather data in operational models. Rainfall not only directly affects payload but also interacts with other operational factors, amplifying its impact.

In summary, both the XGBoost and LSTM models have demonstrated interesting capabilities in forecasting mining payload, with their performance being significantly enhanced by the integration of rainfall data. The median absolute error provides a clear metric for comparison, revealing that while XGBoost offers a slightly better interpretability and handles non-linear relationships effectively. Most importantly, our findings show that incorporating operational planning data—such as the number of trucks scheduled for the next shift—can dramatically reduce major prediction errors. These results lay a strong foundation for future work, where further integration of maintenance planning metrics is expected to drive even greater improvements in forecasting accuracy. Future work should focus on integrating more robust anomaly detection mechanisms and exploring alternative modelling techniques to better address the observed discrepancies.

## CONCLUSION

This study has provided compelling evidence that artificial intelligence models offer a promising pathway to accurately forecast hauling fleet production capacity in open pit mining environments fraught with uncertainty. By fusing operational data, meteorological records, and simulated scenarios, our approach captures the complex, dynamic behaviour of mining operations—enabling more nuanced and proactive short-term planning.

The strategic implications of this work are profound. In an industry where operational disruptions ranging from adverse weather conditions to unforeseen equipment downtimes can lead to significant production setbacks, our predictive framework can become an essential decision-support tool. By quantifying the impact of rainfall, on shift operation performance, the model provides actionable insights that allow planners to adjust fleet sequences in short time. This capability not only improves short-term scheduling but also enhances overall operational resilience by aligning production volumes with budget constraints, stock levels, and project timelines.

Looking ahead, several opportunities exist to further refine and extend the predictive accuracy of our models. One promising avenue is the integration of additional operational constraints that are currently underrepresented. For instance, incorporating the effects of blast events could capture the periodic disruptions that frequently slow down operations. Similarly, embedding absenteeism data, especially regarding operator availability, could provide a deeper understanding of the human factors influencing fleet performance. Moreover, including maintenance scheduling indicators such as MTBF and MTTR can address uncertainties arising from equipment reliability, thereby reducing prediction errors even more.

Beyond these enhancements, the integration of our predictive models with real-time decision support systems offers a strategic advantage. By merging forecast outputs with automated dispatch and scheduling tools, mining companies can transition from reactive management to proactive, data-driven planning. Such integration would allow rapid adjustments in operational sequences, mitigating the impact of unpredictable events and ensuring that resources are optimally allocated across various production scenarios. This shift not only has the potential to boost operational efficiency but also to secure a competitive edge in an increasingly digitised industry.

Furthermore, our work underscores a broader vision for the future of mining operations a vision in which advanced analytics and real-time data converge to form a holistic, adaptive management system. As mining companies continue to embrace digital transformation (Shimaponda-Nawa and Nwaila, 2024), the predictive modelling framework presented here can serve as the cornerstone for a comprehensive strategy that encompasses real-time monitoring, anomaly detection, and automated decision-making. This comprehensive approach promises to enhance both economic performance and safety outcomes by minimising disruptions and optimising resource allocation under uncertain conditions.

In summary, this research lays a foundation for the new era of data-driven, short-term planning in open pit mining. By leveraging the strengths of deep learning models and systematically integrating key environmental and operational variables, our approach transforms forecasting from a reactive exercise into a strategic asset. Future work should explore the synergistic integration of diverse data streams and advanced analytics to further refine these models ensuring that mining operations remain agile, resilient, and strategically positioned to navigate the complexities of modern production environments.

## REFERENCE

Ao, M, Li, C and Yang, S, 2023. Prediction method of truck travel time in open pit mines based on LSTM model, in *Proceedings of the 42nd Chinese Control Conference (CCC)*, pp 8651–8656. https://doi.org/10.23919/CCC58697.2023.10240705

Asif, M, Bessant, J and Francis, D, 2010. Meta-management of integration of management systems, *The TQM Journal*, 22(6):599–613. https://doi.org/10.1108/17542731011085325

Baek, J and Choi, Y, 2020. Deep neural network for predicting ore production by truck-haulage systems in open-pit mines, *Applied Sciences*, 10(5):1657.

Browne, C B, Powley, E, Whitehouse, D, Lucas, S, Cowling, P I, Rohlfshagen, P, Tavener, S, Perez, D, Samothrakis, S and Colton, S, 2012. A survey of Monte Carlo tree search methods, *IEEE Transactions on Computational Intelligence and AI in Games,* 4(1):1–43. https://doi.org/10.1109/TCIAIG.2012.2186810

Cambitsis, A, 2012. A framework to simplify the management of throughput and constraints, Southern African Institute of Mining and Metallurgy, Johannesburg.

Carvalho, M, Sampaio, P and Rebentisch, E, Carvalho, J. Á., & Saraiva, P. (2019). Operational excellence, organisational culture and agility: the missing link? Total Quality Management & Business Excellence, 30(13–14), 1495–1514. https://doi.org/10.1080/14783363.2017.1374833

Chen, T and Guestrin, C, 2016. XGBoost: A scalable tree boosting system, in *Proceedings of the 22nd ACM SIGKDD International Conference on Knowledge Discovery and Data Mining, KDD '16*, pp 785–794. https://doi.org/10.1145/2939672.2939785

Fan, C, Zhang, N, Jiang, B and Liu, W V, 2022. Prediction of truck productivity at mine sites using tree-based ensemble models combined with Gaussian mixture modelling, *International Journal of Mining, Reclamation and Environment*, 37(1):66–86. https://doi.org/10.1080/17480930.2022.2142425

García, S, Luengo, J and Herrera, F, 2015. *Data preprocessing in data mining* (Springer: Cham).

Gonzalez, F R, Raval, S, Taplin, R and Parsons, M B, 2019. Evaluation of impact of potential extreme rainfall events on mining in Peru, *Natural Resources Research*, 28:393–408. https://doi.org/10.1007/s11053-018-9396-1

Hochreiter, S and Schmidhuber, J, 1997. Long short-term memory, *Neural Computation*, 9(8):1735–1780.

Hyndman, R J and Athanasopoulos, G, 2018. *Forecasting: Principles and Practice*, 2nd edn (OTexts). Available from: <https://otexts.com/fpp2/>

Miles, L D, 1961. *Techniques of value analysis and engineering* (McGraw-Hill, New York).

Sánchez, F, Vargas, J, Pereira, J and Baena, L, 2020. Innovation in the mining industry: Technological trends and a case study of the challenges of disruptive innovation, *Resources Policy*, 65:101569. https://doi.org/10.1016/j.resourpol.2020.101569

Shimaponda-Nawa, M and Nwaila, G T, 2024. Integrated and intelligent remote operation centres (I2ROCs): Assessing the human–machine requirements for 21st century mining operations, *Minerals Engineering*, 207:108565. https://doi.org/10.1016/j.mineng.2023.108565

Soofastaei, A, Aminossadati, S, Kizil, M S and Knights, P, 2015. Simulation of payload variance effects on truck bunching to minimise energy consumption and greenhouse gas emissions, in *Proceedings of the Coal Operators' Conference 2015* (eds: N Aziz and B Kininmonth), pp 337–346 (University of Wollongong – Mining Engineering, the Australasian Institute of Mining and Metallurgy – Illawarra, and Mine Managers Association of Australia).

Tlhatlhetji, M and Kolapo, P, 2021. Investigating the effects of rainy season on open cast mining operation: The case of Wescoal Khanyisa Colliery, *Research Square Preprint*, https://doi.org/10.21203/rs.3.rs-870740/v1

Van Houdt, G, Mosquera, C and Napoles, G, 2020. A review on the long short-term memory model, *Artificial Intelligence Review*, 53:5929–5955. https://doi.org/10.1007/s10462-020-09838-1

Wang, Q, Zhang, R, Lv, S and Wang, Y, 2021. Open pit mine truck fuel consumption pattern and application based on multi-dimensional features and XGBoost, *Sustainable Energy Technologies and Assessments*, 43:100977. https://doi.org/10.1016/j.seta.2020.100977

Wang, W, Chakraborty, G and Chakraborty, B, 2021. Predicting the Risk of Chronic Kidney Disease (CKD) Using Machine Learning Algorithm, App*lied* Sci*ence,* 11:202. https://doi.org/10.3390/app11010202